# Should I Stay or Should I Go: Coordinating Biological Needs with ContinuouslyG updated Assessments of the Environment

**Liane M. Gabora**

## ABSTRACT

This paper presents Wanderer, a model of how autonomous adaptive systems coordinate internal biological needs with moment-by-moment assessments of the probabilities of events in the external world. The extent to which Wanderer moves about or explores its environment reflects the relative activations of two competing motivational sub-systems: one represents the need to acquire energy and it excites exploration, and the other represents the need to avoid predators and it inhibits exploration. The environment contains food, predators, and neutral stimuli. Wanderer responds to these events in a way that is adaptive in the short turn, and reassesses the probabilities of these events so that it can modify its long term behaviour appropriately. When food appears, Wanderer be-comes satiated and exploration temporarily decreases. When a predator appears, Wanderer both decreases exploration in the short term, and becomes more "cautious" about exploring in the future. Wanderer also forms associations between neutral features and salient ones (food and predators) when they are present at the same time, and uses these associations to guide its behaviour.

## 1. INTRODUCTION

One approach to modeling animal behaviour is to create an animal that continually assesses its needs, determines which need is most urgent, and implements the behaviour that satisfies that need. However, in the absence of appetitive stimuli such as food or mates, or harmful stimuli such as predators, behaviour is often not directed at the fulfillment of any particular need: an animal either remains still or moves about, and both options haw repercussions on many aspects of survival. So the question is not "What should I do next?", but rather, "Should I stay where I am, conserving energy and minimizing exposure to predators, or should I explore my environment, with the possibility of finding food, mates, or shelter"?

In this paper we present a computational model of how positively or negatively reinforcing stimuli affect an animal's decision whether or not, and if so to what extent, to explore its environment. The model is referred to as *Wanderer*. The architecture of *Wanderer* is an extension of an architecture used to model the mechanisms underlying exploratory behaviour in the absence of positively or negatively reinforcing stimuli (Gabora and Colgan, 1990). The general approach is to consider an autonomous adaptive system as an assemblage of sub-systems specialized to take care of different aspects of survival, and what McFarland (1975) refers to as the "final behavioural common path" is the emergent outcome of the continual process of attempting to mutually satisfy these competing subsystems. The relative impact of each subsystem on behaviour reflects the animal's internal state and its assessment of the dynamic affordance probabilities of the environment (for example, how likely it seems that a predator or food will appear). This distributed approach is similar in spirit to that of Braitenburg (1984), Brooks (1986), Macs (1990) and Beer (1990). We make the simplifying assumption that the only possible beneficial outcome of exploration is finding food, and the only possible harmful outcome is an encounter with a predator. The amount of exploration that Wanderer engages in at any moment reflects the relative activations of a subsystem that represents the need for food, which has an excitatory effect on exploration, and a subsystem that represents the need to avoid predators, which inhibits exploration.

The earlier version of *Wanderer* exhibited the increase and then decrease in activity shown by animals in a novel environment (Welker, 1956; Dember & Earl, 1957; May, 1963: McCall, 1974; Weisler & 1976) and all four characteristics that differentiate the pattern of exploration exhibited by animals raised under different levels of predation were reproduced in the model by changing the initial value of one parameter: the decay on the inhibitory subsystem, which represents the animals assessment of the probability that a predator could appear. In the present paper, we first address how salient events such as the appearance of food or predators affect exploration. *Wanderer* does not have direct access to the probabilities that predators and food will appear but it continually reassesses them based on its experiences, and adjusts its behaviour accordingly. This approach merges Gigerenzer and Murray's (1987) notion of cognition as intuitive statistics with Roitblat's (1987) concept of optimal decision making in animals.

We then examine how initially neutral features can come to excite or inhibit exploration by becoming associated with salient ones (food or predators). It has long been recognized that animals form associations of this kind between simultaneously occurring stimuli or events (Tolman, 1932; Hull, 1943). This is useful since in the real world features are clustered — for example, predators may dwell in a particular type of cavernous rock, so the presence of rocks of that sort can be a useful indicator that a predator is likely to be near. Thus features of environments that contain a lot of food are responded to with increased exploration (even when there is no food in sight) and features of environments that contain many predators inhibit exploration (even when there are no predators).

## 2. ARCHITECTURE OF *WANDERER*

Wanderer consists of two motivational subsystems that receive input from five sensory units and direct their out-put to a motor unit, constructed in Common Lisp (Figure 1). One subsystem represents the need to acquire and maintain energy. It has an excitatory effect on exploration and is linked by a positive weight to the motor unit. Exploration in turn feeds back and inhibits activation of this subsystem: this represents fatigue. Activation of the other subsystem represents the need to avoid predators: it has an inhibitory effect on exploration and is linked by a negative weight to the motor unit. Since every moment that passes without encountering a predator is evidence that there is less need to be cautious, activation of the inhibitory subsystem decreases as a function of time in the absence of predation. In addition, since moving about provides more evidence that there are no predators nearby than does immobility, the inhibitory subsystem, like the excitatory subsystem, receives feedback from the motor unit; its activation decreases by an amount proportional to the amount of exploration that occurred during the previous iteration.

Perception units have binary activations. Activation of unit 0 corresponds to detection of food, activation of unit 1 corresponds to detection of a predator, and activation of units 2, 3, and 4 correspond to detection of rock, tree, and sun respectively. Activation can spread from perception units to subsystems, but not the other way around.

The output for each iteration is either zero, signifying immobility, or a positive number that indicates how much exploration is taking place.

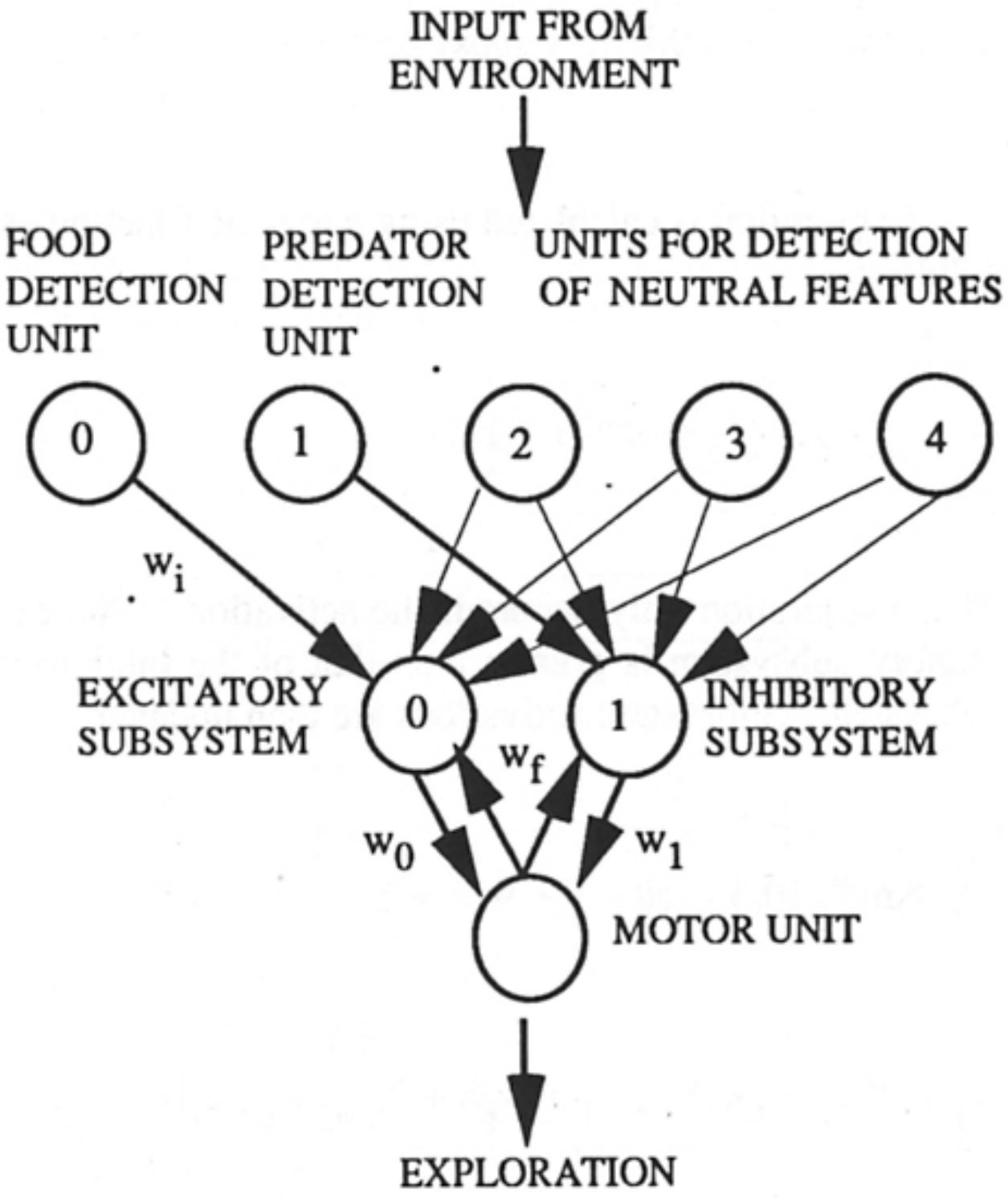

Figure 1. The architecture of Wanderer. Dark lines represent fixed connections. Fine lines represent learnable connections.

The relevant variables and their initial values are

$\mathbf{a_i}$ activation of perception unit $\mathbf{i}$ = {0,11}

**s0** = activation of excitatory subsystem = 0.9

**s1** = activation of inhibitory subsystem = 0.9

**E** = exploration = activation of motor unit

$\mathbf{w_{ij}}$ = weight from perception unit $\mathbf{i}$ to subsystem $\mathbf{j}$:

$\mathbf{w_{0,0}}$ =-0.5, $\mathbf{w_1}$ = 0.9, all others = 0.

$\mathbf{w_0}$ = weight from excitatory subsystem to motor unit

= -0.5

$w_1$ = weight from inhibitory subsystem to motor unit = 0.5

$w_f$ = feedback weight from motor unit to subsystems = -0.1

$k_0$ = rate at which hunger increases= 1.05

$k_1$ = decay on inhibitory subsystem = 0.5

Exploration is calculated using a logistic function as follows:

$$\mathbf{E} = 1 / [1 + e^{-(w0s0 + w1s1)}]$$

Thus exploration only occurs if the activation of the ex-citatory subsystem is greater than that of the inhibitory subsystem. Subsystem activations are then updated:

$$s0_t = \max \{0, k_0 [s0_{t-1} + w_f E + \sum_{i=0}^{n-1} w_{i,0}\, a_i]\}$$

$$s1_t = \max \{0, k_1 [s1_{t-1} + w_f E + \sum_{i=0}^{n-1} w_{i,1}\, a_i]\}$$

## 3. WANDERER'S ENVIRONMENT

Wanderer's environment contains three kinds of stimuli: food, predators, and features that have no direct effect on survival, which will be referred to as neutral features. The initial presence or absence of neutral features is random. The more exploration Wanderer engages in, the greater the probability that a neutral feature will change from present to absent or vice versa in the next iteration:

$c_1$ = constant = 0.75

$p(\Delta a)t = c_1 E_{t-1}$

Perception unit 2 detects a stimulus that is predictive of the appearance of food. Let us say that Wanderer's primary source of food is a plant that grows on a certain kind of soil, and that a certain kind of tree also grows only in that soil, so that the presence oil that tree is predictive of finding food. Perception unit 2 can only turn on when perception unit 0 is on (that is, food can only be detected when the tree is detected). Also, in accord with the harsh realities of life, Wanderer has to explore if it is to find food. The probability of finding food is proportional to the amount of exploration that took place during the previous iteration:

$a_0$ -- activation of food detection unit

$c_2$ = constant specified at run-time

$p(a_0 = 1)_t = c_2 a_2 E_{t-1}$

Perception unit 3 detects a stimulus that is predictive of the appearance of a predator. Let us say that the animal that preys upon Wanderer lives in cavernous rocks, and this unit turns on when rocks of that sort are detected. Perception unit 3 can only turn on when perception unit 1 is on (that is, a predator will only appear when the rock is present). Since predators can appear even when Wanderer is immobile, it is not necessary that Wanderer explore in order to come across a predator.

$a_1$ = activation of predator detection unit

c3 = constant specified at run-time

$p(a_1 = 1)_t = c_3 a_3$

Perception unit 4 detects the presence of the sun, such as when it comes out from under a cloud. The sun is predictive of neither food nor predator.

## 4. EFFECT OF SALIENT STIMULI ON MOTIVATION

### 4.1 IMPLEMENTATION OF SATIETY

Detection of food is represented by the activation of a single binary unit. Activation of this unit decreases the activation of the excitatory subsystem, which in turn brings a short-term decrease in exploration. This corresponds to satiety; once food has been found, the immediate need for food decreases, thus exploration should decrease.

### 4.2 IMPLEMENTATION OF CAUTION

Detection of a predator is represented by the activation of a single binary node that is positively linked to the inhibitory subsystem. Activation of this unit has two ef-fects. First, it causes an increase in the activation of the inhibitory subsystem, which results in a pronounced short-term decrease in exploration. Second, it decreases the decay on the inhibitory subsystem. This has the long-term effect of decreasing the rate at which exploration is disinhibited; in other words, every encounter with a predator causes Wanderer to be more "cautious". Decay on the inhibitory subsystem is updated each iteration as follows:

$\partial = 0.2$

$k1_{min} = 0.5$

$k1_t = \max \{ k1_{min}, (k1_{t-1} + \partial[s1 - s1_{t-1}]) \}$

Since activation of the inhibitory subsystem increases in response to predation, decay increases if a predator appears, and decreases when no predator is present.

## 5. LEARNING ALGORITHM

If a neutral feature — rock, tree, or sun — is present when a salient feature — food or predator — appears, an association forms between the neutral feature and subsystem that is positively linked with the salient feature. Weights on the lines between neutral features and hidden nodes are initialized to zero, corresponding to the state in which no associations, either positive or negative, have formed. If food or a predator appears in the environment, and if one or more initially-neutral features ($p_i$) is present, weights on links connecting initially-neutral features to subsystems are updated as follows:

$\eta$ = learning rate = 0.05

$w_t = w_{t-1} + \eta |s_t - s_{t-1}| a_i$

## 6. RESULTS

Figure 2 plots exploration during a run in which the probability of finding food is high and the probability of predation is zero. Exploration increases quickly initially as activation of the excitatory subsystem increases and the activation of the inhibitory subsystem decreases. It falls sharply whenever food is encountered, and then gradually increases again. Each time food is encountered exploration falls to the same level.

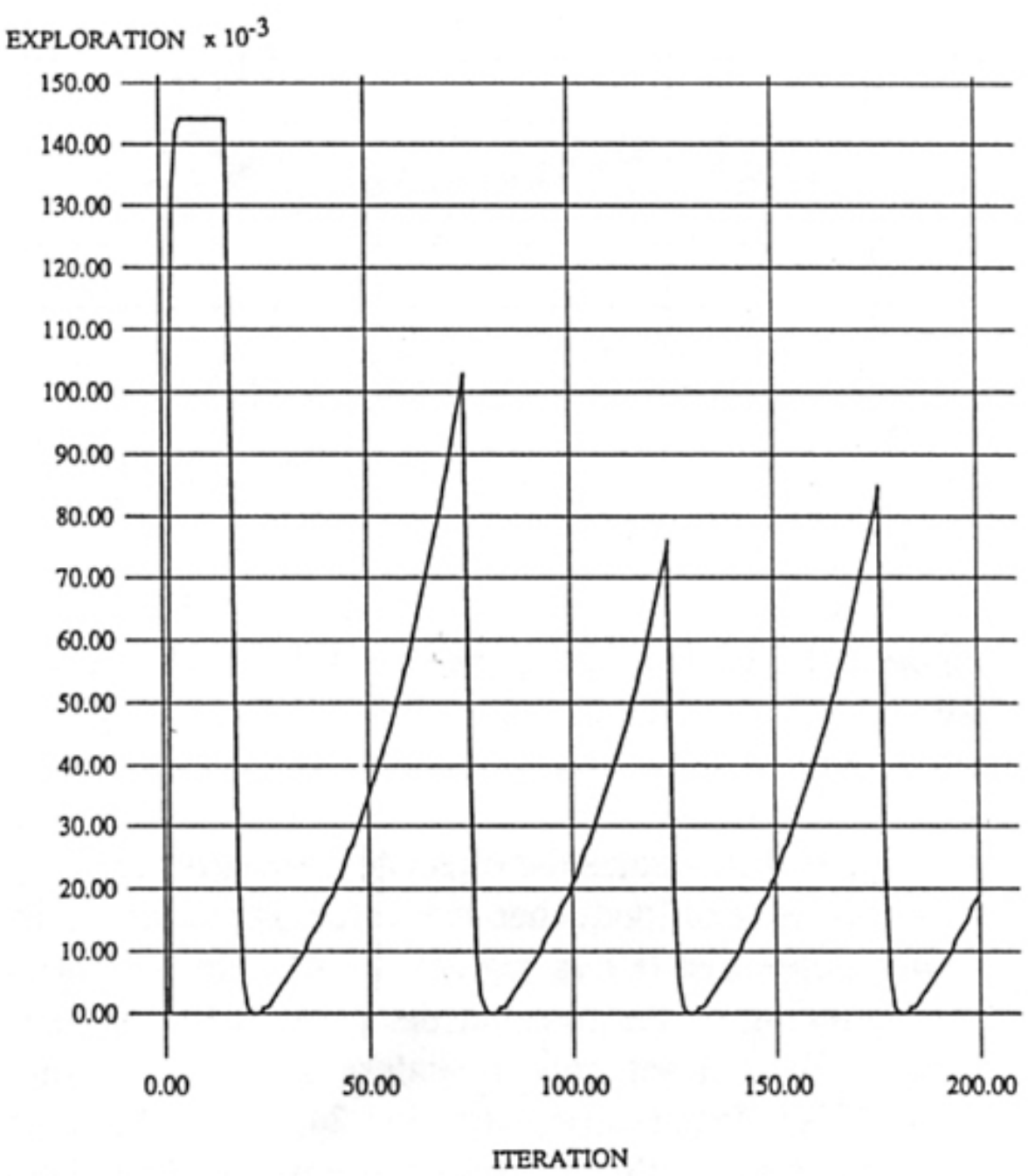

Figure 2. Exploration when p(food) is high and p(predator) = 0.0.

Figure 3 plots exploration during a run in which the probability of finding food is zero and the probability of predation is high. The appearance of a predator causes activation of the inhibitory subsystem to increase, temporarily decreasing exploration. Since no food is present, activation of the excitatory subsystem is high, and exploration quickly resumes. Two more predators are encountered in quick succession. With each encounter, the response is greater, representing an increase in the assessed probability of predation in the current location. Exploration ceases after the third encounter.

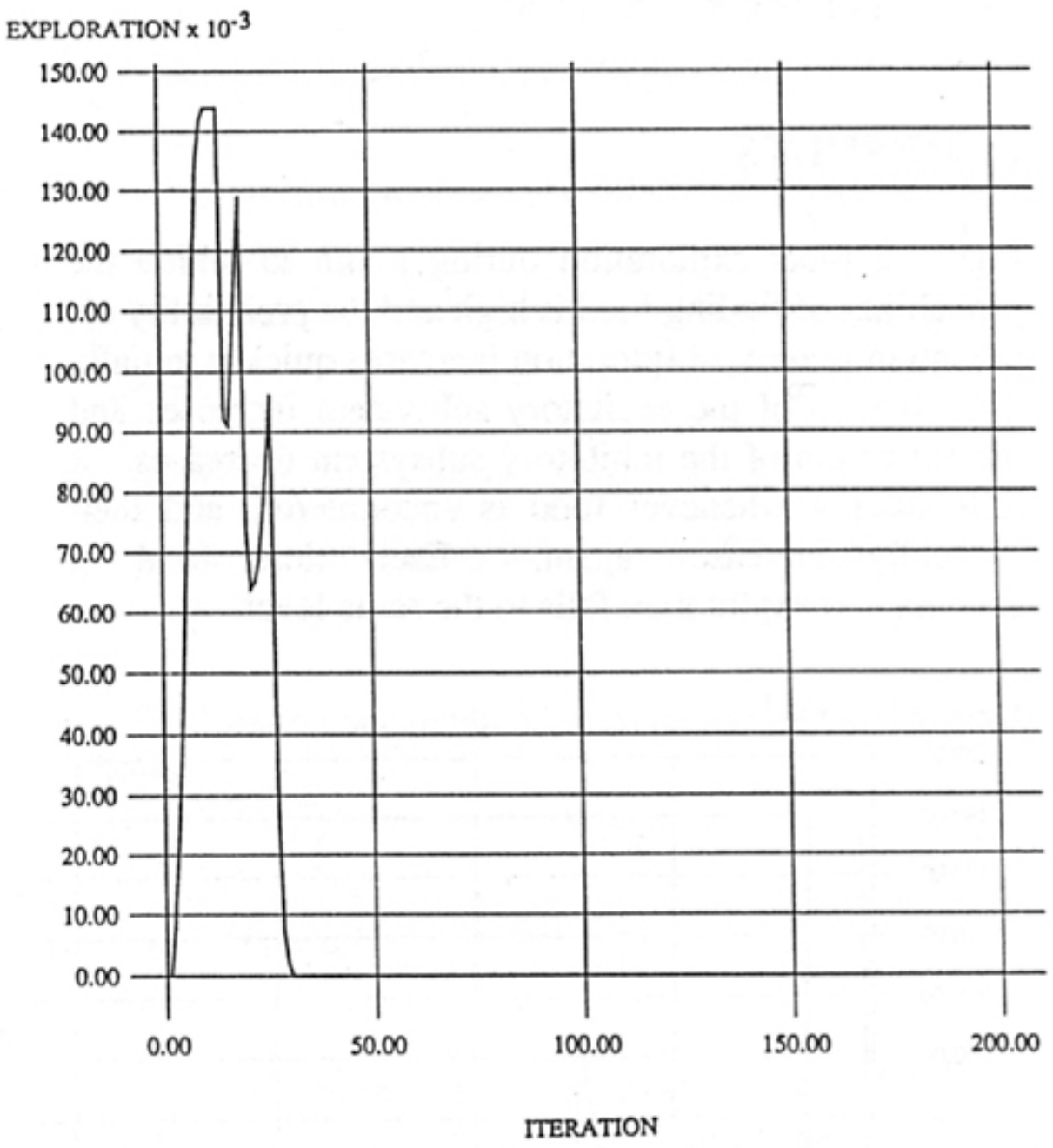

Figure 3. Exploration when p(predator) is high and p(food) = 0.0.

Figure 4 illustrates the effect of associative learning in the presence of food when there are no predators. The exploration curve is less regular. Rocks are present at the beginning of the run, but disappear before food is found. Food never appears unless a tree is present. Food is first found during iteration 24, and exploration drops sharply. At this point associations are formed be-tween food and both the tree and the sun, and the weights on the lines from feature detection units 3 and 4 to subsystem 1 increase (from 0.0 to 0.015). Note that the, association between sun and food is spurious: the sun is not actually predictive of food. Exploration increases until it reaches a plateau. It drops sharply when food is encountered during iteration 80 since at the same time one of the cues predictive of food, the sun, disappears. (Not only is it no longer hungry, but there is indication that there is no food around anyway.) During iteration 84, the tree, the other feature that has been associated with food, disappears as well. Thus exploration increases very slowly. Since little exploration is taking place, features of the environment change little. Rocks appear in iteration 155, but since no association has been formed between rocks and food, this has no effect on exploration. Exploration increases sharply for a brief period between iterations 176 and 183 when the sun comes out, and then again at iteration 193 when trees appear. It plummets once again in iteration 195, with the final appearance of food.

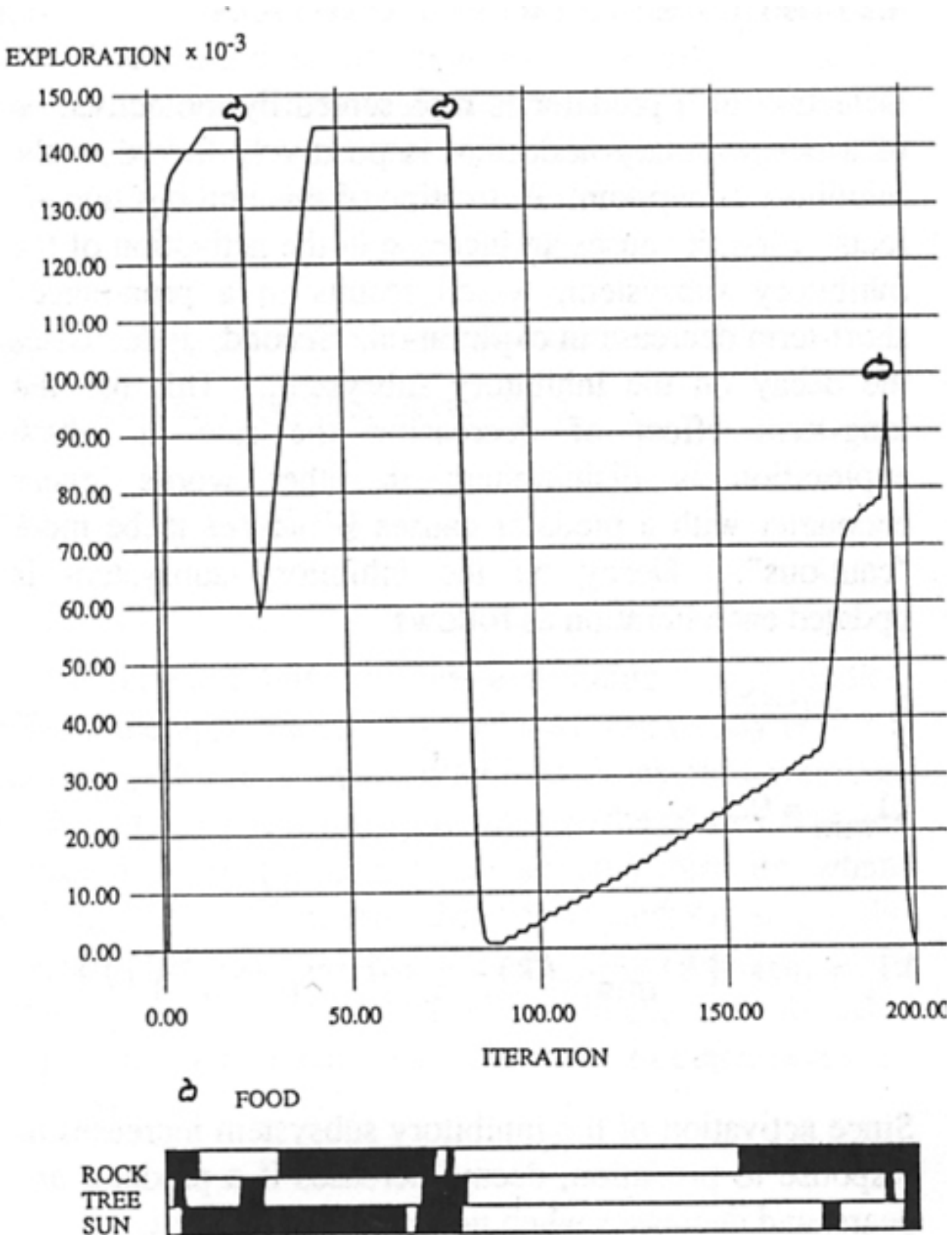

Figure 4. Effect of associative learning when p(food) is high and p(predator) = 0.0. Below: Black bar indicates presence of neutral feature; white bar indicates absence.

The effect of associative learning on response to predation is illustrated in Figure 5. (In this experiment, decay on the inhibitory subsystem is held constant so that exploration does not fall quickly to zero despite the high predation rate.) Response ED predation grows increasingly variable throughout the run, reflecting the ex-tent to which features that have become associated with predation are present at the Lime a predator appears. Wanderer eventually associates all three features of its environment with predators, and none with food. Since two of the three features are present, exploration stops at iteration 179 and does not resume by the 200th iteration. Since Wanderer is not moving, there is no further change in the neutral features until the end of the run.

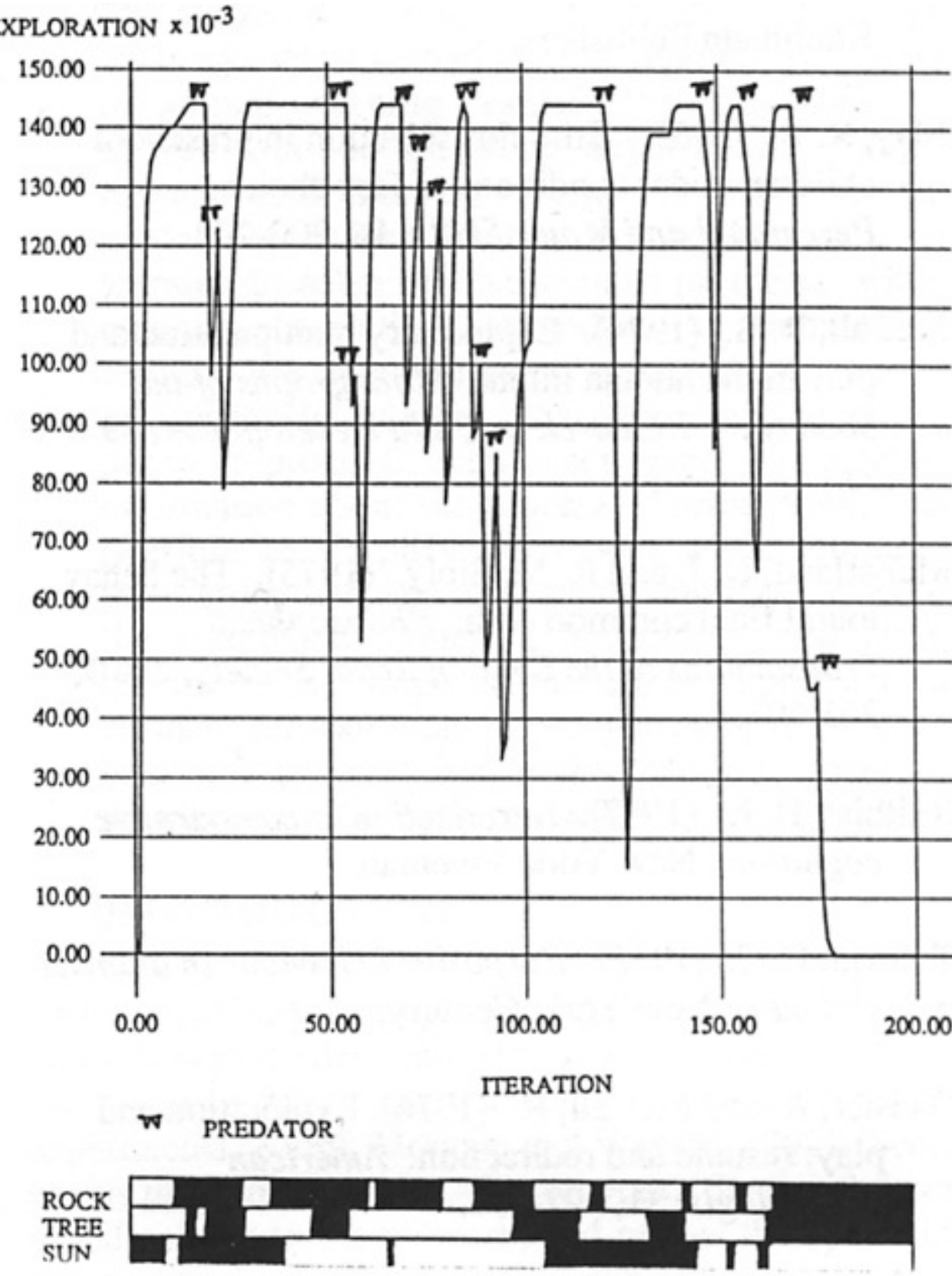

Figure 5. Effect of associative learning when p(food) is high and p(predator) = 0.0. Below: Black bar indicates presence of neutral feature; white bar indicates absence.

The effect of associative learning in the presence of both food and predators is illustrated in Figure 6. Rocks become associated with predators, and trees become associated with food, as expected. (After 200 iterations the weight on the line from Perception Unit 2 to Sub-system 0 is 0.091, and the weight on the line from Perception Unit 3 to Subsystem 1 is 0.170.) However, spurious associations also form between neutral features and salient features (with weights on the relevant lines ranging from 0.046 to 0.112).

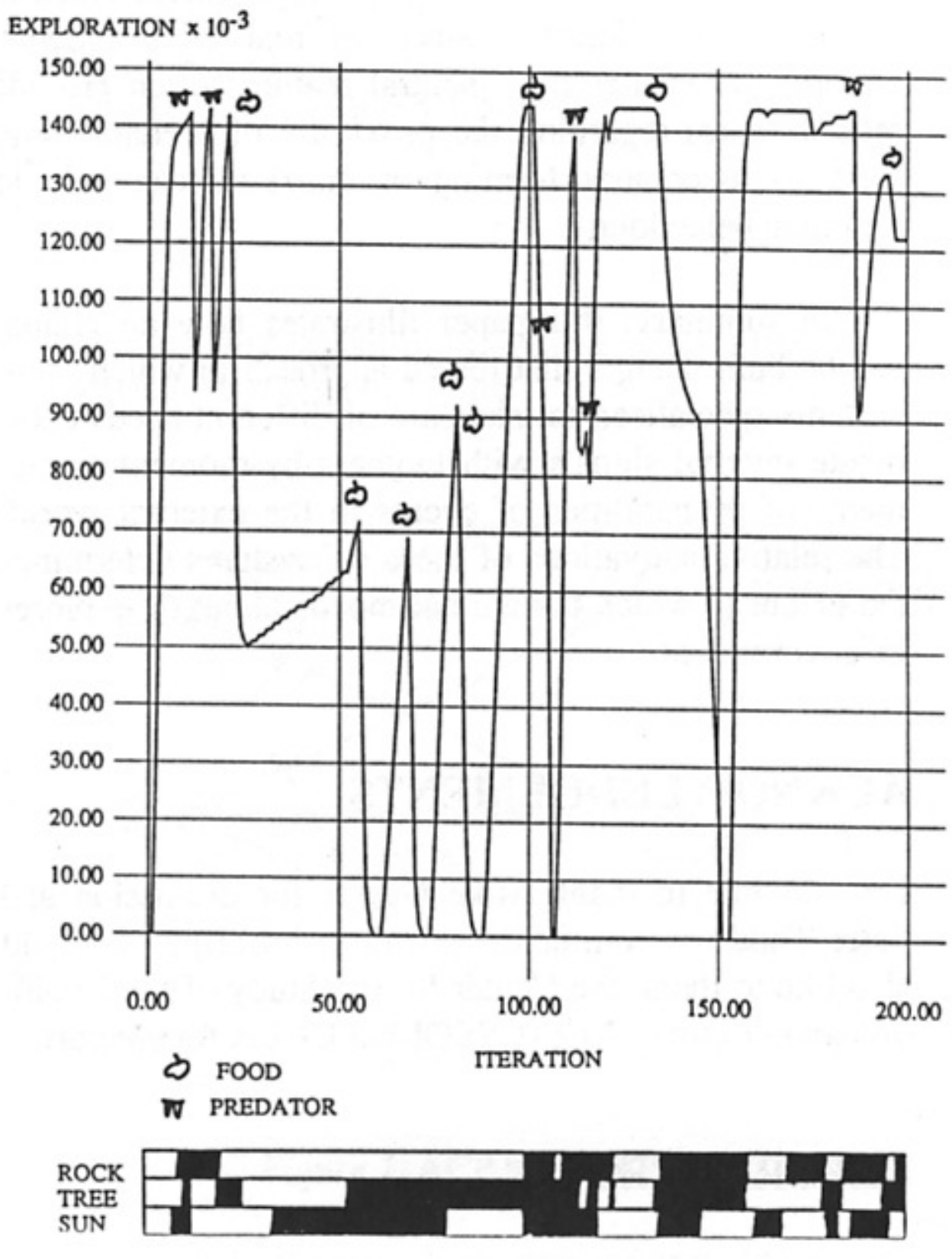

Figure 6. Effect of associative learning when both p(food) and p(predator) are high. Below: Black bar indicates presence of neutral feature; white bar indicates absence.

## 7. DISCUSSION

Wanderer is a simple qualitative model of the mechanisms underlying how animals coordinate internal needs with external affordances. It does not ad-dress a number of real world complexities: the perceptual inputs are ungrounded, and the problems associated with actively moving about in a real environment are bypassed. Since weights never decrease using the delta ride, once associations are formed, they cannot be unlearned. However Wanderer responds to events in a way that is adaptive in the short term, and reassesses the probabilities of these events so that it can modify its long term behaviour appropriately. When food appears, Wanderer becomes satiated and temporarily decreases exploration. When a predator appears, Wanderer both temporarily decreases exploration to avoid being caught, and becomes more cautious in the near future. When predators are not encountered, Wanderer becomes less cautious. Wanderer also forms associations between neutral features of its environment and salient features (predators and food). Since in real environments, features are clustered — neutral features often provide reliable clues regarding the proximity of predators and food — association-forming of this kind can help to optimize behaviour.

In summary, this paper illustrates how an animal can be built using a distributed approach in which sub-systems specialized to take care of different needs coordinate internal signals with moment-by-moment assessments of probabilities of events in the external world. The relative activations of these subsystems determine the extent to which the animal moves about or explores its environment.

## ACKNOWLEDGEMENTS

I would like to thank Mike Gasser for discussion and Peter Todd for comments on the manuscript. I would also like to thank the Center for the Study of the Evolution and Origin of Life (CSEOL) at UCLA for support.